\documentclass[conference]{IEEEtran}

\usepackage{amsmath,amssymb,amsfonts}
\usepackage{algorithmic}
\usepackage{textcomp}
\usepackage{xcolor}

\usepackage{hyperref}

\usepackage[sorting=none]{biblatex}

\usepackage{graphicx}

\begin{document}

	% TEMPLATE
% ------------------------------------------------------------------------------
% \title{Conference Paper Title*\\ 
% 
% 	{\footnotesize \textsuperscript{*}Note: Sub-titles are not captured in
% 	Xplore and should not be used} \thanks{Identify applicable funding
% 	agency here.  If none, delete this.}
% }

\title{
	Spiker: an FPGA-optimized Hardware accelerator for Spiking Neural Networks
}

\author{
		\IEEEauthorblockN{
			Alessio Carpegna
		}

		\IEEEauthorblockA{
			\textit{Control and Computer Eng. Dep.} \\ 
			\textit{Politecnico di Torino}\\ 
			Torino, Italy \\ 
			alessio.carpegna@polito.it
		} 
	\and

		\IEEEauthorblockN{
			Alessandro Savino
		}

		\IEEEauthorblockA{
			\textit{Control and Computer Eng. Dep.} \\
			\textit{Politecnico di Torino}\\
			Torino, Italy \\
			alessandro.savino@polito.it
		}

	\and
	
		\IEEEauthorblockN{
			Stefano Di Carlo
		}

		\IEEEauthorblockA{
			\textit{Control and Computer Eng. Dep.} \\
			\textit{Politecnico di Torino}\\
			Torino, Italy \\
			stefano.dicarlo@polito.it
		}
}

	\maketitle

	\begin{abstract} 

	Spiking Neural Networks (SNN) are an emerging type of biologically
	plausible and efficient Artificial Neural Network (ANN). This work
	presents the development of a hardware accelerator for a SNN for
	high-performance inference, targeting a Xilinx Artix-7 Field
	Programmable Gate Array (FPGA). The model used inside the neuron is the
	Leaky Integrate and Fire (LIF). The execution is clock-driven, meaning
	that the internal state of the neuron is updated at every clock cycle,
	even in absence of spikes.  The inference capabilities of the
	accelerator are evaluated using the MINST dataset. The training is
	performed offline on a full precision model. The results show a good
	improvement in performance if compared with the state-of-the-art
	accelerators, requiring $215\mu s$ per image. The energy consumption is
	slightly higher than the most optimized design, with an average value of
	$13mJ$ per image. The test design consists of a single layer of
	four-hundred neurons and uses around 40\% of the available resources on
	the FPGA. This makes it suitable for a time-constrained application at
	the edge, leaving space for other acceleration tasks on the FPGA.

\end{abstract}


	\section{Introduction}

	%Artificial Neural Networks (ANNs) are one of today’s essential
	%technologies in many applications, like portable electronics, robotics
	%\cite{lu2017industry}, healthcare \cite{shahid2019applications}, and
	%emerging domains such as autonomous driving \cite{luckow2016deep}. 
	Artificial Neural Networks (ANNs) are complex computational learning
	models that often exploit the computing power of enterprise data centers
	and public cloud infrastructures to speed up training and inference.
	However, the cloud-based ANN model is showing its limitations
	\cite{marchisio2019deep}. Moving a large amount of data through the
	Internet implies a considerable energy overhead spread along the
	communication channel. Communication means a non-negligible latency, not
	compatible with performance-constrained applications (e.g., real-time
	systems). This may create a severe bottleneck with the increasing
	pervasiveness of the Internet of Things (IoT) that risks saturating the
	network communication towards centralized servers
	\cite{shafique2018overview}. Finally, there are applications in which,
	for security reasons, data must be kept local.

	All these factors push toward moving part of the ANN processing, if not
	all, towards the edge of the network, closer to where the data
	originated and where responses are required. However, when ANNs are
	deployed at the edge, they cannot feature the same computing power as in
	the cloud, where high-performance GPUs and computing cores are
	available. %ANNs are composed of many neurons that work in parallel, and
	%they poorly fit a general-purpose processor, which executes
	%instructions sequentially or with a limited degree of parallelism. 
	Specialized accelerators can cover this gap to speed up ANN inferencing
	at the edge \cite{chen2014diannao}. Nevertheless, hardware acceleration
	alone is not enough to fully benefit from the power of ANNs at the edge.
	Learning models must be adapted to this new limited environment. 	
	
	In Convolutional Neural Networks (CNNs), one of the dominant ANN models,
	each neuron requires many calculations (mainly multiplying and
	accumulating) at every cycle. This creates a pattern suitable for
	massively parallel implementations but wastes resources. Moreover, if
	the goal of ANNs is to mimic the behavior of the human brain, this model
	is far from its biological counterpart. Spiking Neural Networks (SNN),
	an emerging type of event-based neural network, can make a difference in
	this sense \cite{putra2020fspinn}. 
	%Conversely, SNN reduce wasted
	%computation by only processing received events, i.e., neurons are
	%updated only when specific events arrive, mimicking the biological
	%neuron's behavior.
	In an SNN, the information between neurons is exchanged in form of
	binary spikes, thus minimizing resources to link neurons in the network.
	Additionally, time is treated as an additional dimension in the input
	and this makes SNNs more suitable for processing time series. 
	
	In the past, SNNs were mainly implemented on CPUs using software
	frameworks such as Brian/Brian2 \cite{brian2} or in GPU-based frameworks
	such as CARLsim4 \cite{chou2018carlsim}. Since these computing
	architectures cannot efficiently support the sparse feature of SNNs,
	specific ASIC processors have been proposed, e.g., TrueNorth from IBM
	\cite{akopyan2015truenorth} or SpiNNaker \cite{painkras2012spinnaker}.
	When hardware resources are limited, Field Programmable Gate Arrays
	(FPGAs) offer a great technology to implement several accelerators on
	the same hardware block to speed up different tasks thanks to their
	in-field programmability~\cite{spiker, minitaur,wangQian2017Eepn,darwin,sixuLi,
	frenkelCharlotte2019A021}.  However, the existing designs still face a
	complexity that may prevent the deployment of extensive networks at the
	edge.

	This paper presents Spiker, a new FPGA-based SNN hardware accelerator
	with off-line training based on the well-known Leaky Integrate and Fire
	(LIF) neuron model. Spiker aims to reduce at a minimum the size of the
	architecture, with the specific goal of maximizing the allowed
	parallelism, reducing the required execution time to fit
	performance-constrained applications at the edge.  This is obtained by
	introducing well-crafted approximations, reducing the size of the
	neuron. The target dataset used to evaluate the designed accelerator is
	the MNIST~\cite{deng2012mnist}. The training is performed offline with a
	full precision model, using the Spike Timing Dependent Plasticity
	(STDP) unsupervised learning method \cite{stdp}. In contrast, the
	inference is performed on the accelerator with the simplified neuron
	structure.  Experimental results show that this simplified model
	introduces a slight accuracy loss compared to the full precision
	counterpart, provides high throughput, and maintains low hardware
	complexity.
	
	%The remaining of the paper is organized as follows:
	%\autoref{sec:background} overviews the basic SNN concepts required to
	%understand the behavior of the accelerator. Section
	%\ref{sec:architecure} introduces the designed architecture and the
	%steps followed to make it as light as possible. Section
	%\ref{sec:results} experimentally evaluates the capability of the
	%proposed accelerator and \autoref{sec:conclusions} summarizes the main
	%contributions of the paper and overviews future activities.

	\section{Background}
\label{sec:background}

\begin{figure}[ht]
			\centering
			\includegraphics[width=0.8\columnwidth]
			{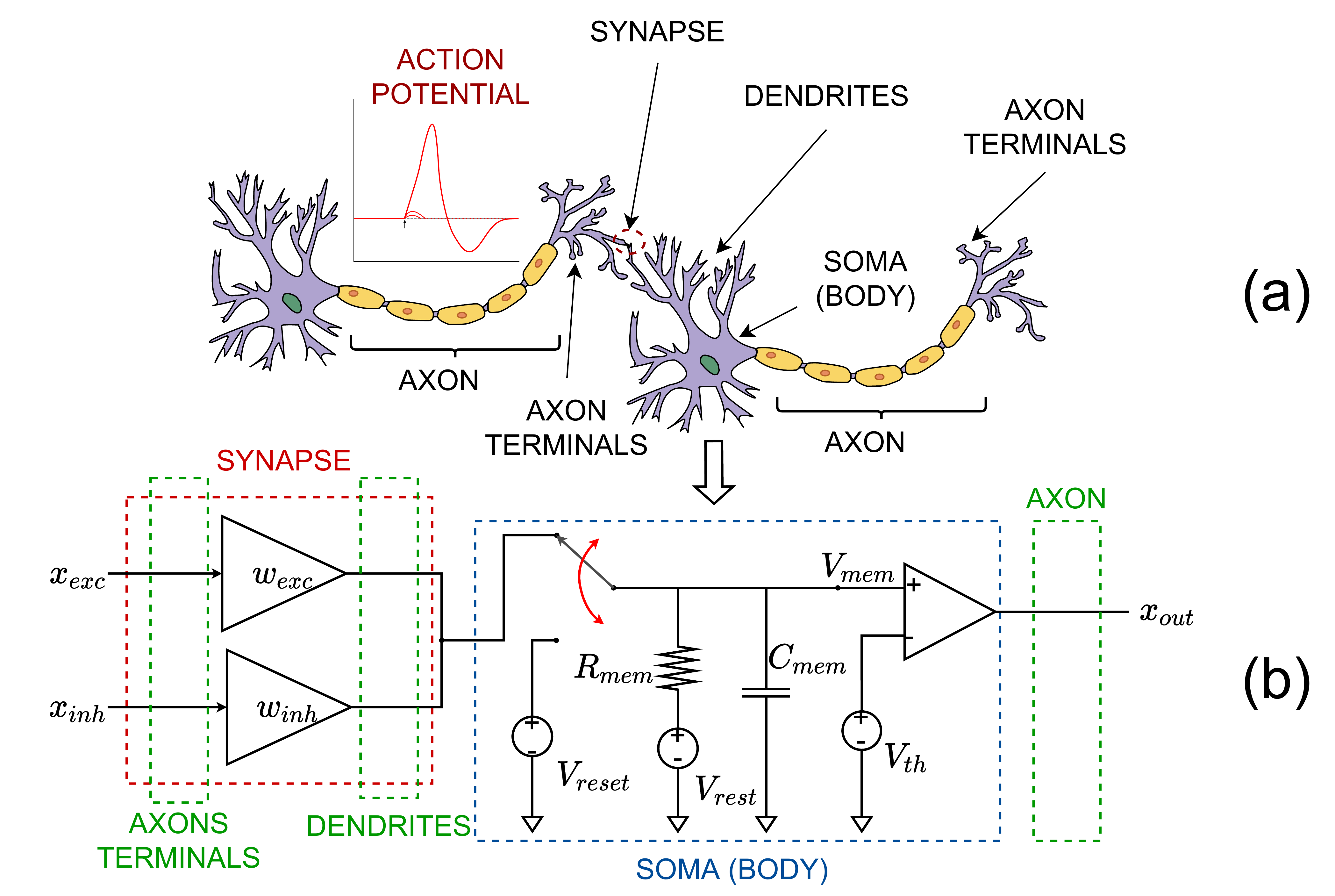}
			\caption{SNN neuron concept: (a) biological structure,
			(b) equivalent circuit of the LIF model.}
			\label{fig:neuronModel}
		\end{figure}

	Spiking neural networks have first emerged in computational neuroscience
	to model the behavior of biological neurons. The body of a neuron
	(\autoref{fig:neuronModel}-a), also called soma, is characterized by an
	internal state associated with a voltage across its membrane (i.e.,
	membrane potential). Stimuli (i.e., signal spikes) received at the input
	terminals of the neuron (i.e., dendrites) can modify the membrane
	potential. In particular excitatory stimuli cause an increment in the
	membrane potential, while inhibitory stimuli decrease it. If the
	potential exceeds a threshold, an action potential takes place. The
	action potential is a sudden increase in the membrane voltage, which
	then rapidly tends towards its rest value~\cite{actionPotential}. This
	generates a voltage spike propagating to other neurons through the
	output terminals (i.e., axons).

	Many mathematical models have been developed in the last decade to
	quantitatively describe this kind of behavior. 	The Hodgkin-Huxley model
	is the most accurate and precisely describes the behavior of ions within
	the membrane \cite{hodgkinHuxley}. It is very realistic but too complex
	for hardware implementations. The Izhikevich model exploits mathematical
	properties to simplify the Hodgkin-Huxley model \cite{izhikevich}, and
	some works use this model in hardware accelerators
	\cite{kitCheung2009Apsn}. However, our architecture aims to reduce the
	area occupation, and the Izhikevich model is too complex to accomplish
	this goal.  The simplest available model is the Integrate and Fire (IF)
	model, which treats the membrane as an ideal capacitance. However, this
	model is too approximated and leads to behaviors quite different from
	those observed in a biological neuron.  Eventually, the Leaky Integrate
	and Fire (LIF) model is a good trade-off between simplicity and
	biological realism. 
	
	The LIF model treats the membrane as a leaky capacitance, including a
	resistive part that forces the voltage towards a rest value in the
	absence of input stimuli. The temporal evolution of the membrane
	potential can be described through the characteristic equations of the
	LIF equivalent circuit (\autoref{fig:neuronModel}-b).
	\autoref{eq:lifDifferentialEquation} governs the membrane potential
	$V(t)$, where $V_{rest}$ is the resting potential in the absence of
	stimuli, $t_0$ is the initial time (e.g., the instant in which the
	neuron receives a spike), and $\tau$ is the time constant of the
	equivalent RC-circuit, namely $\tau = R_{mem} \cdot C_{mem}$.

		{\footnotesize
		\begin{equation}
			\frac{dV(t)}{dt} = \frac{1}{\tau} \cdot (V_{rest} -
				V(t_0))
			\label{eq:lifDifferentialEquation}
		\end{equation}}
	Solving the differential equation leads to an exponential evolution of
	the membrane potential in the form of:

		{\footnotesize\begin{equation}
			V(t) = V_{rest} + (V(t_0) - V_{rest}) \cdot
				e^{-\frac{t-t_0}{\tau}}
			\label{eq:lifExponential}
		\end{equation}}
	A specific conductance acting as a weight characterizes each synapse,
	i.e., the interface between the dendrites and axons of two neurons.  The
	conductance modifies the incoming spikes, leading to a variation in the
	membrane potential proportional to the synapse's weight.  This
	conductance-based model is complex despite being a faithful emulation of
	biological behavior.  A simpler alternative is the current-based synapse
	model. In this case, the synapse is treated as a simple charge
	amplifier. The input arrives in the form of an ideal current spike that
	affects the membrane potential proportionally to the charge it delivers. 

	Spike-timing-dependent plasticity (STDP) is among the most used learning
	algorithms in SNNs \cite{stdp}. This biologically inspired learning rule
	modifies the synaptic strength (i.e., weights) as a function of the
	relative timing of pre- and post-synaptic spikes. The details of the
	learning procedure are not reported in this work since the accelerator
	has been designed for inference. The user is free to decide the training
	method enabling reaching the desired goal, such as accuracy or
	biological plausibility.

	\section{Architecture}
\label{sec:architecure}

\begin{figure}[ht]
				\centering
				\includegraphics[width=0.95\columnwidth]
				{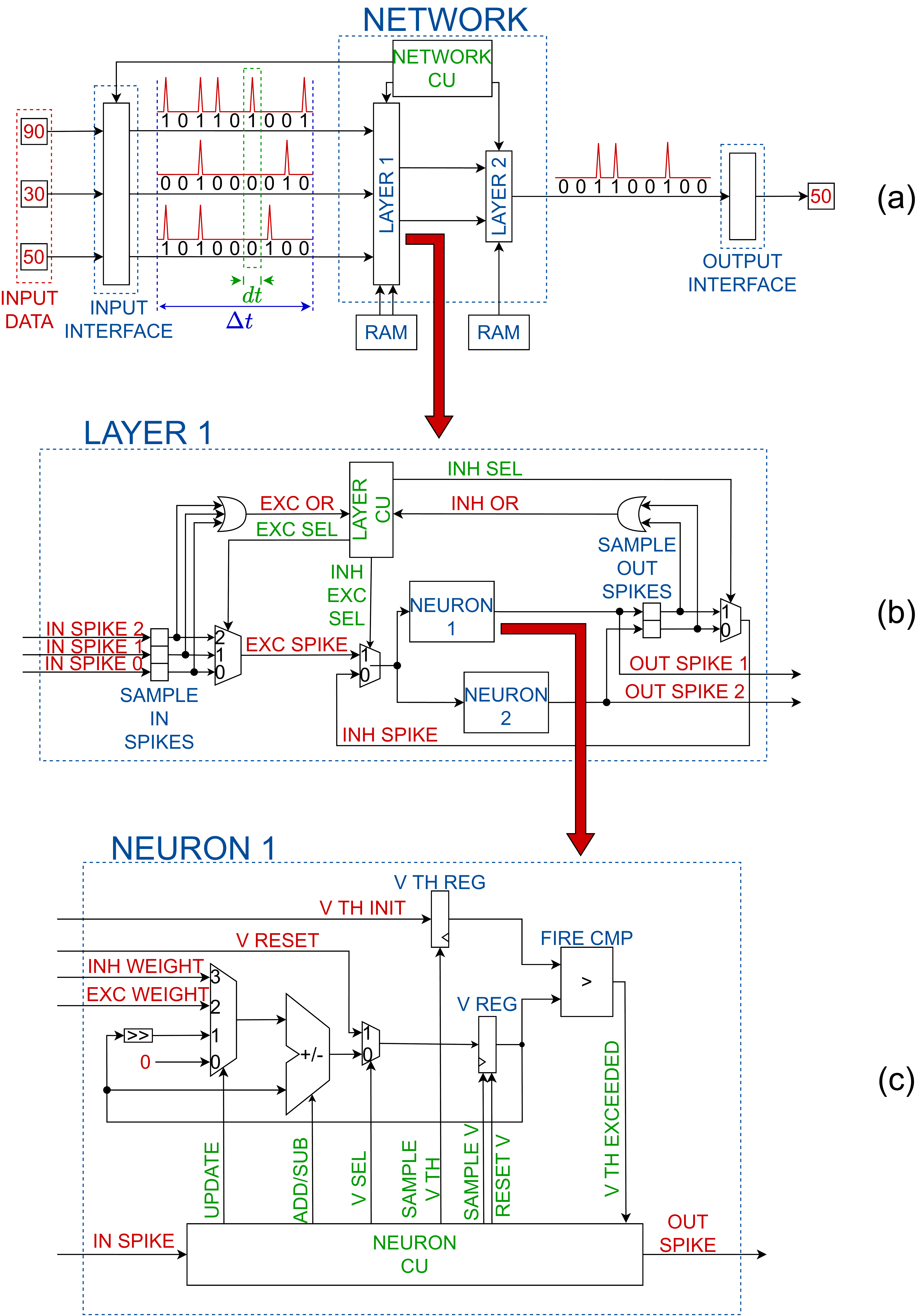}
				\caption{Network architecture}
				\label{fig:networkArchitecture}
			\end{figure}

	This section overviews the general architecture of Spiker (\autoref{fig:networkArchitecture}). 
	It focuses on the optimizations introduced to reduce the hardware complexity and improve performance.
	%Without losing in generality, the architecture is presented using a running example based on 
	%digits recognition, i.e., inputs of the SNN are images of digits to be recognized. 
	Overall, Spiker includes an input and an output layer used to interface 
	the spiking core of the network with the external data and several hidden 
	layers processing spikes. 
	Spiker implements a clock-driven neuron architecture, i.e., in the absence of spikes,
	the membrane potential is updated at every clock cycle 
	following the exponential trend reported in~\autoref{eq:lifExponential}.
	However, inputs are processed only when at least one spike is present at the input of a layer.
	This solution is more power-hungry than pure event-driven architectures 
	but allows reducing the required hardware resources to a minimum. 
	Finally, the architecture works with fixed-point arithmetic.

	\subsection{Input interface}
	\label{subsec:interface}

	SNNs process numerical data vectors that must be converted into sequences of spikes. Spikes are represented as single bits in the digital domain to minimize resources. There
	are different methods available for this conversion, depending on the
	type of input data \cite{8689349}: (i) firing rate coding (i.e., information is 
	encoded using the instantaneous average firing rate), (ii) population rank coding 
	(i.e., information is encoded using the relative firing time of a population
	of neurons), or (iii) temporal coding (i.e., information is encoded with the exact timing of individual spikes). 
	Temporal encoding is the most biologically realistic encoding suited for dynamically evolving data.
	Nevertheless, Spiker uses firing rate coding that works well for static data such as images and enables compact and optimized hardware implementation. 
	
	In Spiker, one input data (e.g., the intensity of a pixel in an image) is treated as an instantaneous firing rate (i.e., the probability that a spike occurs within an interval). The conversion process starts selecting the duration of the spike sequence ($\Delta t$ in \autoref{fig:networkArchitecture}-a) that is the same for all inputs. The sequence is split into computation steps of duration $dt$, each able to accommodate a single spike. This parameter defines the temporal resolution of the network. The product of the numerical input ($rate$) by $dt$ provides the average number of spikes $ASPS = total\ spikes / total\ steps = rate \cdot dt$ that must be generated in each step of the whole spike sequence. $ASPS$ is a number between zero and one since one spike per step can occur at most. The conversion process finally requires generating a random number $n$ between zero and one. If $n$ is greater than $ASPS$, a spike takes place. This method allows generating a random sequence of statistically independent spikes whose timing follows a Poisson distribution \cite{poisson}.

This approach requires the generation of random values both during training and inference. During training (performed off-line), a statistically independent random value is generated for each network input. This guarantees high accuracy, leading the network to generalize the input patterns better. However, Spiker uses a single random value for all input data in parallel once the network has been trained. This strongly reduces the hardware cost with a negligible effect on the accuracy during inference. The generation of a pseudo-random value on-board uses a Linear Feedback Shift Register (LFSR). The substantial area reduction introduced by the use of a single LFSR for all inputs allows to increase the period of the generated random sequence, equal to $2^{bit-width}$ in a maximal LFSR, thus improving the quality of the random numbers with a negligible impact on the global area.

	\subsection{Output interface}

		The output interface translates the sequences of spikes generated at the network's output into numerical information that can be further processed. Spiker implements this interface using simple counters, one for each output neuron. The value of these counters should be normalized by the duration of the spike sequence ($\Delta t$) to obtain the firing rate of the output neurons. However, being $\Delta t$ the same for all neurons, this operation can be avoided, thus saving area. The computed firing rates can then be used to infer, for example, by looking at which neuron has been the most active for a specific input pattern.

	\subsection{Network architecture}

		The network can be composed of an arbitrary amount of layers,
		connected in a feed-forward structure (\autoref{fig:networkArchitecture}a). A central control unit (CU)
		manages the elaboration that is organized into temporal steps. 
		When all the layers are ready, the CU orders the input interface 
		to generate a new set of spikes, corresponding to one
		temporal step $dt$. The spikes are generated in parallel for all
		the inputs. Then the CU enables all layers in parallel to start a new
		elaboration on the inputs. The first layer takes directly the
		input spikes. All the others use the output generated by the
		corresponding previous layer as an input.  In this way the
		information is propagated in the feed-forward direction in a sort of
		pipeline.

		When a layer ends its computation, it informs the central CU that it is ready and then waits. Since 
		elaboration in different layers can take different time, the main CU waits for all layers to finish and then starts a new elaboration cycle. This process is repeated until the entire sequence of input spikes is elaborated.

	\subsection{Layer}

		A layer (\autoref{fig:networkArchitecture}-b) can be composed of an arbitrary number of neurons,
		updated in parallel. A dedicated layer CU manages the elaboration.
		In particular, in Spiker, each neuron can elaborate one spike at a
		time. The first role of the layer CU is to provide the input spikes
		one-by-one to all the neurons in parallel.

		Spiker gives the possibility to implement inter-layer inhibitory connections. Each neuron can be connected to all the other neurons of the layer
		using links with negative weights. This creates an inhibitory effect that reduces their membrane potential,
		preventing a neuron from firing a spike. 
		The layer control unit also manages the elaboration of the spikes coming from inhibitory
		connections. The process is the same seen for the excitatory
		connections. The spikes are sampled in parallel, and then, once the
		elaboration of the excitatory spikes ends, the control unit switches
		to the inhibitory ones providing them one-by-one to all
		the neurons.

		Since spikes are elaborated one by one, a complete update cycle can take considerable time. One of the
		advantages of the SNNs is that, if the model and the input encoding are well designed, the spike sequences received and
		generated are pretty sparse. 
		To evaluate this characteristic, a statistical analysis is performed considering the MNIST dataset used in \autoref{sec:results} in our experimental setup. The goal is to verify how many steps $dt$ within the sequence $\Delta t$ contain at least one active spike. 

			\begin{figure}[ht]
				\centering
				\includegraphics[width=0.8\columnwidth]
					{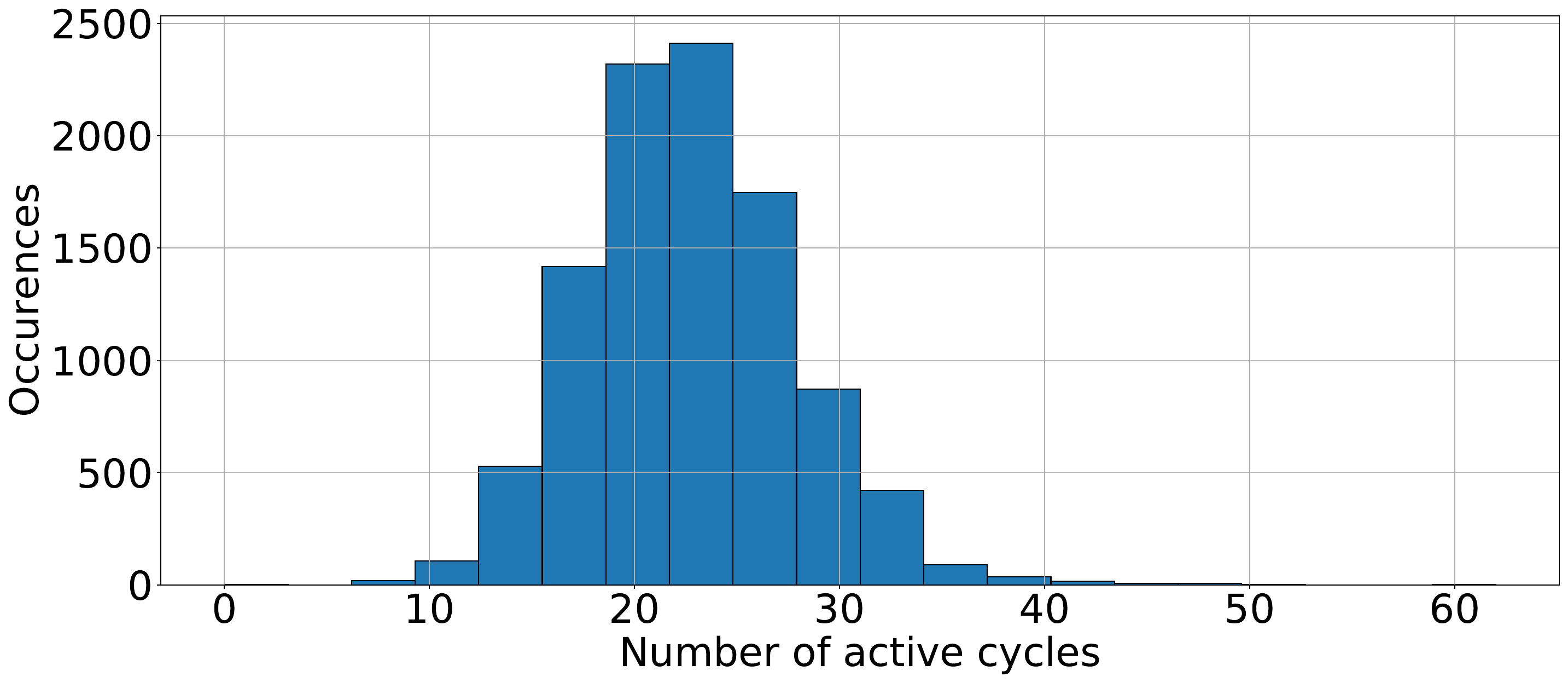}
				\caption{Statistics of the active elaboration
					steps (i.e., cycles with at least one
					spike) on all images of the MNIST
					dataset. Obtained values have been
					computed with a window of 3,500
					elaboration steps.}
				\label{fig:activeCycles}
			\end{figure}

		\autoref{fig:activeCycles} shows that, on average, twenty-three steps are active on
		a total of 3,500, which corresponds to less than the 1\%. A
		similar result has been obtained for the inhibitory spikes.
		Spiker uses this characteristic to improve its performance. In all steps without active
		spikes, performing computations would be a waste of time and power. A simple OR operation
		applied on all spikes (both excitatory and inhibitory) enables Spiker to skip the elaboration in all steps without active spikes.

	\subsection{Neuron \label{sec:veuron}}

		The architecture of the neuron is minimal
		(\autoref{fig:networkArchitecture}-c). This is the strong
		point of the designed accelerator. Being the architecture
		clock-driven, the exponential decay, corresponding to
		\autoref{eq:lifDifferentialEquation}, is solved as:

			\begin{equation}
				V[n] = V[n-1] + \frac{dt}{\tau} \cdot (V_{rest}
					- V[n-1])
				\label{fig:clockDrivenExp}
			\end{equation}

		Here is where the two main optimizations are performed. First, the internal voltage parameters, that are the rest
		potential, the reset potential and the threshold, are moved,
		using an offset, in order to obtain $V_{rest} = 0V$. This
		simplifies \autoref{fig:clockDrivenExp} to:

			\begin{equation}
				V[n] = V[n-1] - \frac{dt}{\tau} \cdot V[n-1]
				\label{fig:simplifiedClockDrivenExp}
			\end{equation}
		thus removing the need to add $V_{rest}$, reducing the required
		computations by one. See \autoref{sec:results} for a practical
		example.

		Second, the equation still involves a multiplication
		($\frac{dt}{\tau} \cdot V[n-1]$), which requires a lot of 
		hardware resources. To avoid it the quantity $\frac{dt}{\tau}$ is
		approximated to the nearest negative power of two. In this way
		the multiplication can be substituted with a simple bit shift,
		which can be computed with zero cost in terms of
		hardware components. The advantage of such an approximation is
		that, if the model is designed properly, so choosing $dt$ and
		$\tau$ such to make their ratio equal or near to a negative
		power of two already in the training phase, the approximation has
		no, or at least a small, impact on the accuracy.

		The other operations that the neuron can perform are: (i) resetting
		the membrane potential to $V_{reset}$ when it exceeds the
		threshold, forcing it to $V_{rest} = 0V$ (through the synchronous
		reset signal RESET V) at the end of the complete elaboration
		in order to start all the elaborations from the same state, and (ii)
		add an excitatory or inhibitory weight to the membrane potential
		when an active spike is received.

		Finally, one additional advantage of the designed structure is
		that it allows initializing the threshold to an arbitrary
		value, which can be different for all the neurons. This provides the
		freedom to use it as an additional hyperparameter that can be tuned 
		for each neuron during the learning phase.

	\subsection{Weights memory}

		The memory required by the weights is generally too big to fit
		into the flip-flops available in an average FPGA. For this reason, Spiker stores the weights in an external
		memory. In this case, the access parallelism of the
		memory becomes the main performance bottleneck since it
		determines how many weights can be accessed, and so how many
		neurons can be updated in parallel. Many FPGAs, like the one
		used in this paper (see \autoref{sec:results}),
		are equipped with Block RAM (BRAM). A BRAM is a memory,
		divided into blocks that are directly integrated within the FPGA.
		The blocks can be accessed in parallel, enabling a high
		degree of parallelism in the accelerator. Quantization of the
		weights is performed to reduce the required bit-width
		to a minimum, thus reducing the memory occupation and increasing the
		number of weights that can be accessed in parallel with the same
		memory bandwidth.

		While processing the spikes sequentially, the layer provides the
		index of the currently analyzed spike as an output. A dedicated circuit is then required to translate such an index to the physical address for all the BRAMs that must be accessed in parallel.

	\section{Experimental results \label{sec:results}}

	\subsection{Experimental set-up}
	\label{subsec:experimentalSetup}

		Spiker has been tested on the MNIST dataset, 
		a database of 28x28 grey-scale images with eight-bit
		per pixel. The SNN network developed by Diehl and Cook \cite{peterDiehl} is taken as a reference. 		
		Table \ref{tab:referenceModel} summarizes
		the original parameters of the model and the values used in Spiker after shifting
		$V_{rest}$ to 0mV (i.e., a shift of 65.0mV) as proposed in \autoref{sec:veuron}. The evaluation of the accelerator is performed using the whole test set, composed of 10,000 images.
		
			\begin{table}[htb]
                \caption{Optimized model parameters}
				%\caption{Reference model parameters for the evaluation and values after moving $V_{rest}$ to 0. The inhibitory weight ($w_{inh}$) is the same for all the 	connections.}

				\label{tab:referenceModel}

				\centering

				\begin{tabular}{|c|c|c|}

					\hline

					\textbf{Parameter}		& 
					\textbf{Original Value}		& 
					\textbf{Used Value}		\\

					\hline

					$V_{rest}$			&
					-65.0mV				&
					0mV				\\

					\hline

					$V_{reset}$			&
					-60.0mV				&
					5.0mV				\\

					\hline

					$V_{th0}$			&
					-52.0mV				&
					13.0mV				\\

					\hline

					$\tau$			&
					100ms				&
					-				\\

					\hline

					$w_{inh}$			&
					-15				&
					-15				\\

					\hline

					$\Delta t$			&
					350ms				&
					350ms				\\

					\hline

					$\delta t$			&
					0.1ms				&
					0.1ms				\\
					
					\hline

					Cycles				&
					3500				&
					3500				\\

					\hline

					$\frac{\delta t}{\tau}$		&
					$\frac{0.1ms}{100ms} = 10^{-4}$	&
					$2^{-10}$			\\

					\hline

				\end{tabular}
				
			\end{table}

		According to the reference model, the parameters are tuned for a
		network structure with 400 neurons and inhibitory connections
		between the	neurons. This structure allows reaching
		acceptable accuracy results, even if non-optimal, as seen in the
		dedicated section, with a total number of neurons reasonable to
		integrate on an average FPGA. It is important to remark here
		that the goal of the evaluation is not to show the superiority
		of this model compared to other neural networks (e.g., CNNs).
		This model was selected since it is used in other studies
		proposing FPGA-based SNN hardware acceleration and allows us to
		compare Spikers with its competitors.
		
		The network works with fixed-point precision. The bit-width used
		inside the neuron is set to 16 bit (3 fractional and 13 integer).
		Instead, the quantization of weights is down to 5 bit (3 fractional
		and 2 integers). The total memory required is therefore $400
		\times 28 \times  28 \times 5$ bit = $196$ KByte. The training
		is performed in full-precision using STDP on a custom python
		simulation of the model.

	\subsection{Accuracy results}
	
		\autoref{tab:results} shows the effect of the neuron approximations introduced in Spiker on the accuracy of the network. The training accuracy is reported only for the
		reference and the Spiker LIF full precision models, since
		the training is always performed in full precision. The
		various simplifications are then applied only during inference, for
		which the accelerator is designed. \autoref{tab:results}
		shows that:

			\begin{enumerate}

				\item Changing the model from a
					conductance-based \cite{peterDiehl}
					solution to a lighter current-based
					alternative has no significant impact on
					the training phase and implies a
					relatively small 1.42\% reduction during
					inference.

				\item Using a single random number generator in
					the input interface,
					(\emph{Current-based 1 LFSR} in
					\autoref{tab:results}), causes an
					additional 2.19\% accuracy reduction.

				\item Working with an internal
					parallelism of 16 bits and quantizing
					the weights down to a 5 bits width
					implies a further 1.01\% accuracy
					decrease.

				\item The approximation of the ration between
					$\delta t$ and $\tau$ has no effect on
					the accuracy, since the starting value
					was quite near to $2^{-10}$.

			\end{enumerate}

			\begin{table}[htb]

				\caption{Training and test results with
					different simplifications}

				\label{tab:results}

				\centering

				\begin{tabular}{|c|c|c|}

					\hline

					\textbf{Model}			& 
					\textbf{Training}		& 
					\textbf{Inference}		\\

					\hline

					Peter Diehl\cite{peterDiehl}	&
					80.54\%				&
					78.58\%				\\

					\hline

					Spiker LIF model full precision	&
					80.22\%				&
					77.16\%				\\

					\hline

					Spiker LIF model 1 LFSR		&
					-				&
					74.97\%				\\

					\hline

					Spiker LIF model 16bit/5bit	&
					-				&
					73.96\%				\\

					\hline

				\end{tabular}
				
			\end{table}

		So the overall accuracy reduction of the completely simplified
		model, when compared to the reference one, is around 4.5\%,
		which is more than acceptable considering the substantial
		hardware simplification.

	\subsection{Area and performance results}

		The designed accelerator has been synthesized on a medium-size
		Xilinx Artix-7 FPGA. All the four-hundred neurons are
		instantiated as physical components and can be updated in
		parallel. The available BRAM allows accessing all the
		required weights in parallel. The routing gives no
		issues.

			\begin{table}[htb]

				\caption{Required area}

				\label{tab:areaPerformance}

				\centering

				\begin{tabular}{|c|c|c|c|}

					\hline

					\textbf{HW component}		& 
					\textbf{LUT}			& 
					\textbf{FF}			&
					\textbf{BRAM}			\\

					\hline

					Single neuron			&
					62				&
					40				&
					-				\\

					\hline

					400 neurons			&
					23885				&
					15614				&
					-				\\

					\hline

					Complete layer			&
					26038				&
					16846				&
					-				\\

					\hline

					Spikes generator		&
					6343				&
					6311				&
					-				\\

					\hline

					Weights				&
					-				&
					-				&
					45				\\

					\hline
					\hline

					Complete accelerator		&
					29145 (55\%) 			&
					26853 (25\%)			&
					45 (32\%)			\\

					\hline

					Total FPGA Available		&
					53200				&
					106400				&
					140				\\

					\hline

				\end{tabular}
				
			\end{table}

		\autoref{tab:areaPerformance} shows the required hardware
		resources. The last row summarizes all the available components.
		Overall, the FPGA usage is around 55\% for the LUTs, 25\% for
		the FFs and 32\% for the available memory. This is a
		significant result considering the high number of instantiated
		neurons.

	\subsection{Comparison with other accelerators}

			\begin{table*}[tb]

				\caption{Comparison table}

				\label{tab:comparisonTable}

				\centering

				\begin{tabular}{|c|c|c|c|c|c|}

					\hline

					\textbf{Design}			& 
					\cite{minitaur}			&
					\cite{wangQian2017Eepn}		&
					\cite{darwin}			&
					\cite{sixuLi}			&
					This work			\\

					\hline

					\textbf{Clock frequency(MHz)}	& 
					75				&
					120				&
					25				&
					100				&
					100				\\

					\hline

					\textbf{Data format}		&
					16bit Fixed			&
					8bit Fixed			&
					32bit Fixed			&
					16bit Floating		&
					16bit Fixed			\\

					\hline

					\textbf{Computing scheme}	&
					Event-Driven			&
					Clock-Driven			&
					Event-Driven			&
					Adaptive Clock/Event-Driven	&
					Clock-Driven			\\

					\hline

					\textbf{Neuron model}		&
					LIF				&
					LIF				&
					LIF				&
					LIF				&
					LIF				\\

					\hline

					\textbf{FPGA platform}		&
					Spartan 6			&
					Virtex 6			&
					Spartan 6			&
					Virtex 7			&
					Artix 7				\\

					\hline

					\textbf{Neurons}		&
					1794				&
					1591				&
					1794				&
					1094				&
					1384				\\

					\hline

					\textbf{Synapses}		&
					647000				&
					638208				&
					647000				&
					177800				&
					313600				\\

					\hline

					\textbf{Task}			&
					MNIST				&
					MNIST				&
					MNIST				&
					MNIST				&
					MNIST				\\

					\hline

					\textbf{Computation time}	&
					$0.53s$/image			&
					$8.40s$/image			&
					$0.16s$/image			&
					$3.15ms$/image			&
					$215\mu s$/image		\\

					\hline
					
					\textbf{Computation time @100MHz}	&
					$0.40s$/image			&
					$10.08s$/image			&
					$40.00ms$/image			&
					$3.15ms$/image			&
					$215\mu s$/image		\\

					\hline

					\textbf{Energy}			&
					$0.80J$/image			&
					$1.12J$/image			&
					Not reported			&
					$5.04mJ$/image			&
					$13mJ$/image			\\

					\hline

					\textbf{Energy/Synapse}		&
					$1.2\mu J$/synapse		&
					$1.76\mu J$/synapse		&
					Not reported			&
					$0.028\mu J$/synapse		&
					$0.041\mu J$/synapse		\\
					
					\hline

				\end{tabular}
				
			\end{table*}

		\autoref{tab:comparisonTable} shows the comparison between
		different accelerators tested on MNIST. The table reports the key results to highlight the contributions of this work and to compare it to other similar works. Comparison is performed with accelerators based on the same neuron model (LIF) and targeting the same dataset (MNIST). 
		For a fair comparison of the performance, the time required to elaborate an image is normalized, considering a working frequency of 100MHz for all the accelerators.
		The reported computation time for Spiker ($215\mu s$) is the
		fastest among the considered accelerators, demonstrating how all
		the design optimizations described before make the processing
		structures required for the computation simpler and more
		efficient. This result is also significant considering that the
		number of synapses that can be instantiated in the fastest
		competitor is half the number instantiated with Spiker. From the
		energy standpoint, while the reported energy per image is twice
		the best design, i.e., \cite{sixuLi}, the energy per synapse
		makes the energy consumption almost comparable. Nevertheless,
		considering that the best competitor in terms of energy
		consumption relies on a mixed strategy that includes an
		event-driven part, this makes the developed architecture
		particularly promising. As a reference, the test of the model
		using Brian 2 \cite{brian2} required 0.2s on a 2GHz Intel i5 dual-core
		processor.

	\section{Conclusions}
\label{sec:conclusions}

	This work presented Spiker, a small high-performance SNN hardware
	accelerator synthesized on a Xilinx Artix-7 FPGA and tested on the
	MNIST dataset. Spiker significantly accelerates 
	inference, with a competitive energy consumption
	and a limited impact on the accuracy. Presented results are a good starting point 	for future work, in which different and deeper network structures will
	be considered. The main goal will be to improve the classification
	accuracy, making it comparable with other state-of-the-art accelerators.
	This, together with its competitive performance, can make Spiker a very
	good solution for performance-constrained applications at the edge.

	\printbibliography

\end{document}